\documentclass{article}
\usepackage{spconf,amsmath,graphicx,hyperref}

\title{FlowIID: Single-Step Intrinsic Image Decomposition via Latent Flow Matching}
\name{Mithlesh Singla, Seema Kumari, and Shanmuganathan Raman}
\address{Indian Institute of Technology Gandhinagar, India\\ 
\{mithlesh.singla, seema.kumari, shanmuga\}{@iitgn.ac.in}}

\begin{document}
%
\maketitle
\begin{abstract}

Intrinsic Image Decomposition (IID) separates an image into albedo and shading components. It is a core step in many real-world applications, such as relighting and material editing. Existing IID models achieve good results, but often use a large number of parameters. This makes them costly to combine with other models in real-world settings. To address this problem, we propose a flow matching-based solution. For this, we design a novel architecture, FlowIID, based on latent flow matching. FlowIID combines a VAE-guided latent space with a flow matching module, enabling a stable decomposition of albedo and shading. FlowIID is not only parameter-efficient, but also produces results in a single inference step. Despite its compact design, FlowIID delivers competitive and superior results compared to existing models across various benchmarks. This makes it well-suited for deployment in resource-constrained and real-time vision applications.

\end{abstract}
\begin{keywords}
Intrinsic image decomposition, albedo, shading, latent flow matching, and single-step generation. 
\end{keywords}

 \section{Introduction}
\label{sec:intro}
Intrinsic Image Decomposition (IID) aims to separate an input image $I$ into its reflectance (albedo, $A$) and shading ($S$) under the Lambertian assumption~\cite{barrow1978recovering}: $I = A \cdot S$. Albedo encodes material and color, while shading captures shape and illumination. IID supports tasks including relighting, material editing, object detection, and recognition, thus requiring both accuracy and efficiency in practice.

Existing IID methods adopt two paradigms: (i) separate estimation of $A$ and $S$~\cite{luo2020niid,cgintrinsics}, which often results in reconstruction inconsistencies; and (ii) direct shading prediction with albedo computed as $A = I/S$~\cite{careagaIntrinsic,fan2018revisiting}, which aligns the reconstruction and is popular in recent approaches.


Conventional IID models rely on deep convolutional networks that are computationally expensive and difficult to deploy. Diffusion-based methods~\cite{kocsis2024intrinsic,rgbx} achieve strong results but require slow multi-step inference and large parameter counts. To overcome these limitations, we propose a latent flow matching (LFM) framework that learns continuous transport in a compact latent space. Our architecture integrates a UNet with an encoder–decoder design, achieving efficient decomposition with substantially fewer parameters. This approach offers (i) deterministic training compared to stochastic diffusion, (ii) single-step, fast inference, and (iii) competitive accuracy with a compact model size. Our main contributions are as follows:

(1) We present the first application of LFM to intrinsic image decomposition, enabling efficient single-step prediction.

(2) We introduce a VAE-guided latent representation that stabilizes the decomposition process.

(3) We achieve superior performance on standard benchmarks while ensuring practicality for real-time and embedded deployment.
 
\section{Related Work}
\label{sec:related}
Before the deep learning era, intrinsic image decomposition (IID) methods often relied on additional cues such as depth information~\cite{chen2013simp}. However, recent works no longer incorporate such auxiliary information or strong prior assumptions about the scene. With the introduction of large-scale datasets such as CGIntrinsics~\cite{cgintrinsics} and Hypersim~\cite{hypersim}, neural network-based approaches have become the dominant paradigm~\cite{baslamisli2018joint, careagaIntrinsic}.

Current state-of-the-art models typically rely on architectures with a large number of parameters. For instance, Careaga and Aksoy~\cite{careagaIntrinsic} employ a three encoder–decoder design, while Kocsis et al.~\cite{kocsis2024intrinsic} and $RGB\leftrightarrow X$~\cite{rgbx} fine-tune the pre-trained text-conditional Stable Diffusion V2 model~\cite{rombach2022high} by treating intrinsic layers as multichannel images. Similarly, Luo et al.~\cite{luo2024intrinsicdiffusion} leverage a diffusion model combined with ControlNet~\cite{zhang2023adding} to generate intrinsic modalities.

Although these models achieve state-of-the-art performance on benchmarks such as ARAP~\cite{ARAP} and MIT Intrinsic~\cite{grosse2009ground}, they require hundreds of millions of parameters. Since IID often serves as a fundamental preprocessing step for downstream applications such as relighting, material estimation, and LDR-to-HDR conversion, their computational overhead makes integration into broader vision pipelines challenging. To address this limitation, we propose a latent flow matching-based approach with a lightweight architecture that integrates a UNet with an encoder–decoder design.

 \begin{figure*}[t]
  \centering
  \includegraphics[width=\textwidth,  trim={0 0 3.5cm 0},clip]{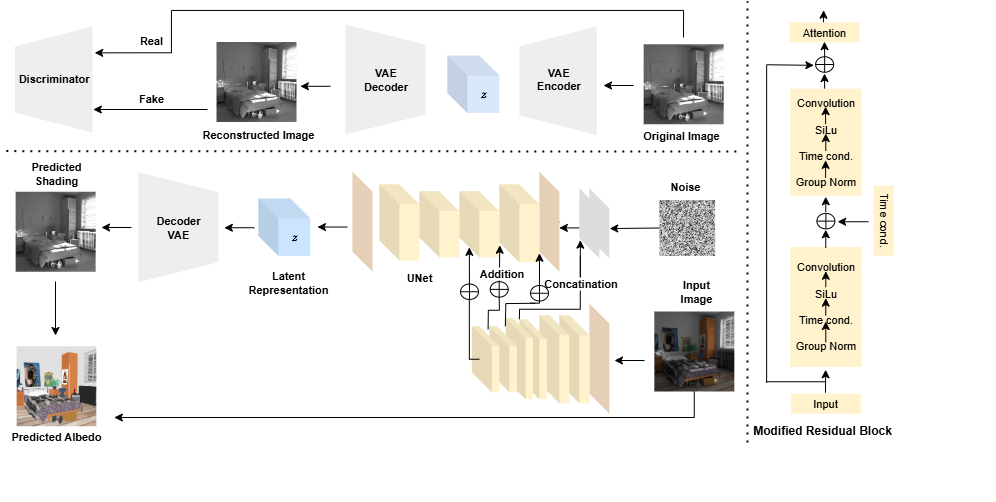}
  \vspace{-1.75cm}
  \caption{Architecture of FlowIID. The upper part illustrates that VAE and a discriminator are trained in a VAEGAN setup, and the lower part illustrates model inference.  The model follows an encoder–decoder design, where the input image and noise are mapped to a latent representation via the encoder and a UNet backbone with modified residual blocks (shown on the right). The VAE decoder generates the shading component from the latent vector, and the albedo is obtained by dividing the input image by its shading. The VAE encoder and discriminator are used only during training and are not required at inference time.}
\label{fig:block_diagram}
\end{figure*}

 \subsection{Background of Flow Matching}
 \label{ssec: Flow Matching}
 Recently, Flow Matching ~\cite{lipman2022flow} has emerged as a powerful technique in generative modeling. It uses vector fields to convert samples from a source distribution $p_0(x_0)$ (usually Gaussian) at time $t=0$ to a complex target distribution $p_1(x_1)$ at time $t=1$ in a high-dimensional space. To achieve this, we define intermediate distributions $p_t(x_t)$ for $t \in [0,1]$, which evolve according to the ordinary differential equation (ODE):
 \begin{equation}
 d x_t = v_t(x_t) \, dt
\label{sol}
 \end{equation}
Let $\theta$ be our model parameters. At any time step $t \in [0,1]$, model outputs $u_\theta(x_t, t)$ and will try to predict the $v_t$. Training is performed by minimizing the mean squared error:
\begin{equation}
\theta = \arg\min_{\theta}{E}_{t, x_t} \left\| u_{\theta}(x_t, t) - v_t \right\|^2
 \label{loss}
 \end{equation}
 
 Let $x_0 \sim p_0$ be a sample from the source distribution and $x_1 \sim p_1$ be a sample from the target distribution. Let $\sigma_{\min}$ denote a minimum scaling factor. During training, we randomly select a time step $t \in [0,1]$ and using optimal transport (OT) equation ~\cite{lipman2022flow} we get conditional path
 \begin{equation}x_t = \big(1 - (1 - \sigma_{\min}) t \big) x_0 + t x_1\end{equation}
 and velocity\begin{equation}v_t = x_1 - (1 - \sigma_{\min}) x_0  \end{equation}
Our model will predict $ u_{\theta}(x_t, t)$ and calculate loss using equation~\eqref{loss}.
 During sampling, we start at the time step $(t=0)$ and go to $(t=1)$ iteratively. At each time step $(t)$ our model output velocity $u_{\theta}(x_t, t)$ and we solve it for probability path using equation~\eqref{sol}

\begin{figure*}[t]
  \centering
  \includegraphics[width=0.9\textwidth]{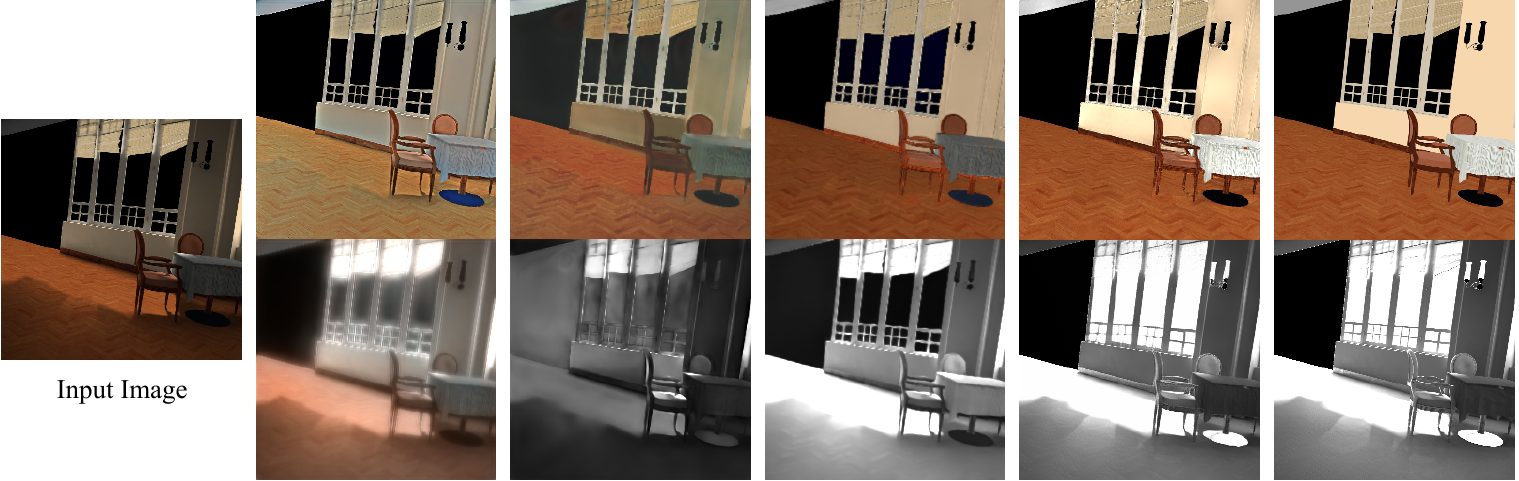}
  \vspace{-0.4cm}
  \caption{Qualitative comparison of proposed model with existing work, from left column - (i) input image, (ii) Lettry et al.~\cite{lettry2018unsupervised}, (iii) Niid-net~\cite{luo2020niid}, (iv) Careaga and Askoy~\cite{careagaIntrinsic}, (v) Ours, (vi) Ground Truth. The figure shows the albedo and shading components predicted by our model alongside the ground truth and prior methods. Our model produces consistent shading while preserving the color fidelity of the albedo image.}
\label{fig:visual_comparison}
\end{figure*}
\section{Proposed Approach}
\label{sec:proposed}
Given an input image $I$, our model decomposes it into shading $S$ and albedo $A$ components according to the relation $I = A \cdot S$. To address the complexity of this task while ensuring resource efficiency, we propose FlowIID, a single-step latent flow matching-based architecture (Figure~\ref{fig:block_diagram}). FlowIID operates in two stages: First, the input image and noise are passed through the encoder and UNet to obtain a compact latent shading representation. Second, this latent representation is passed to the VAE decoder to generate the predicted shading. The albedo is then recovered by dividing the input image by the estimated shading. This design integrates generative modeling with intrinsic decomposition, enabling accurate and stable estimation of scene reflectance and illumination. We now describe the architecture of the proposed model and loss function in detail below.


\textbf{Model Architecture:}
Our proposed model, FlowIID, consists of four main components: a VAE, a discriminator, a UNet, and an encoder. The VAE and discriminator are implemented using lightweight convolutional layers. Given an input image of shape $H \times W$, the VAE produces a latent representation of shape $8 \times H/8 \times W/8$. Inspired by the residual block design in~\cite{residual}, we incorporate a modified residual block (MRB) into both the UNet and the encoder, as illustrated in Fig.~\ref{fig:block_diagram} (right side). The UNet contains two downsampling blocks and two upsampling blocks with skip connections, along with two convolutional projection layers at the ends. Attention layers are included only in the second and third blocks to balance accuracy and efficiency. The UNet takes as input a noise tensor of shape $8 \times H/8 \times W/8$ and predicts a velocity vector of the same dimension.

The encoder contains six downsampling blocks (without attention) and an initial convolutional layer. It processes an input image of shape $3 \times H \times W$. The output from the third block has shape $256 \times H/8 \times W/8$, which is concatenated with the noise tensor of shape $8 \times H/8 \times W/8$. This produces an effective input of shape $264 \times H/8 \times W/8$ for the first convolutional layer of the UNet. Furthermore, the outputs from the last three encoder blocks are pointwise added to the outputs of the UNet’s input convolution layer and its two downsampling blocks, respectively.

\textbf{Training and Loss Functions:}
We train FlowIID in two stages. In the first stage, we train the VAE and a discriminator in a VAEGAN-type setup. Given a shading image $s_0$ of shape $H \times W$, the VAE encoder $E$ produces a latent representation $E(s_0)$ of shape $8 \times H/8 \times W/8$. Then we pass the latent representation $E(s_0)$ through the VAE decoder $D$ to get the reconstructed image $\hat{s}_0$. So, our reconstruction loss $\mathcal{L}_{\text{rec}}$ is 
\begin{equation}
\mathcal{L}_{\text{rec}} = \left\| \hat{s}_0 - s_0 \right\|^2 .
\label{eq:vae_rec}
\end{equation}
We further employ a perceptual loss $\mathcal{L}_{\text{perc}}$ and a KL-divergence loss $\mathcal{L}_{\text{KL}}$ for the latent distribution.

Training for VAE proceeds in two steps. For the first 90 epochs, the objective is
\begin{equation}
\mathcal{L}(E,D) = \mathcal{L}_{\text{rec}} 
+ 0.005 \, \mathcal{L}_{\text{KL}} 
+ \mathcal{L}_{\text{perc}} .
\end{equation}
In the subsequent 200 epochs, we introduce the adversarial loss $\mathcal{L}_{\text{A}}$ with weight $0.1$, giving
\begin{equation}
\mathcal{L}(E,D) = \mathcal{L}_{\text{rec}} 
+ 0.005 \, \mathcal{L}_{\text{KL}} 
+ \mathcal{L}_{\text{perc}} 
+ 0.1 \, \mathcal{L}_{\text{A}} .
\label{eq:vae_total}
\end{equation}
In the second stage, we train our flow matching model.
We use loss function given in~\eqref{loss}, for 250 epochs. Both the VAE and Flow Matching networks are trained with a batch size of 32 and a learning rate of $1 \times 10^{-4}$.

\section{Experimentation}
\label{sec:experimentation}
\textbf{Datasets and Preprocessing}
To train our model, we use three standard datasets: Hypersim~\cite{hypersim}, InteriorVerse~\cite{zhu2022learning}, and the Multi-Illumination Dataset (MID)~\cite{multi_illum}. Hypersim and InteriorVerse provide HDR images with albedo, while MID provides only HDR images, for which we use albedo from Careaga and Askoy~\cite{careagaIntrinsic}. Shading images are obtained by dividing HDR by albedo, then tonemapped to $[0,1]$ without gamma compression~\cite{hypersim}. Multiplying LDR shading with albedo yields white-balanced LDR ground truth images in $[0,1]$. All images are resized to $256 \times 256$ via aspect-ratio preserving scaling and cropping. 
\vspace{-0.5cm}
\begin{table}[ht!]
\centering
\caption{Quantitative comparison of reflectance and shading prediction on MIT Intrinsic dataset}
\resizebox{\columnwidth}{!}{
\begin{tabular}{l ccc ccc}
\hline
Method & \multicolumn{3}{c}{Albedo} & \multicolumn{3}{c}{Shading} \\
\cline{2-4} \cline{5-7}
 & MSE$\downarrow$ & LMSE$\downarrow$ & DSSIM$\downarrow$ & MSE$\downarrow$ & LMSE$\downarrow$ & DSSIM$\downarrow$ \\
\hline
CasQNet~\cite{ma2020casqnet}       & 0.0091 & 0.0212 & 0.0730 & 0.0081 & 0.0192 & 0.0659 \\
PAIDNet.~\cite{huang2025deep}    & 0.0038 & 0.0239 & 0.0368 & \textbf{0.0032} & 0.0267 & \textbf{0.0475} \\
USI3D~\cite{liu2020unsupervised}        & 0.0156 & 0.0640 & 0.1158 & 0.0102 & 0.0474 & 0.1310 \\
Cgintrinsics~\cite{cgintrinsics} & 0.0167 & 0.0319 & 0.1287 & 0.0127 & 0.0211 & 0.1376 \\
PIENet~\cite{das2022pie}       & \textbf{0.0028} & 0.0126 & \textbf{0.0340} & 0.0035 & 0.0203 & 0.0485 \\
\textbf{Ours}               & 0.0040 & \textbf{0.0043} & 0.0435 & 0.0109 & \textbf{0.0119} & 0.0823  \\
\hline
\end{tabular}
}
\label{tab:mit}
\end{table}

For testing, we evaluate FlowIID on two standard benchmarks: ARAP~\cite{ARAP} and the MIT Intrinsic dataset~\cite{grosse2009ground}.  
For ARAP, we follow the protocol of~\cite{careagaColorful}, which removes duplicate scenes and adds three scenes from MIST~\cite{hao2020multi}. Since this protocol introduces some ambiguity in duplicate scene removal and dataset extension, we test baseline models from their publicly available checkpoints to ensure fairness.  
For MIT Intrinsic, we adopt a similar train and test split of Barron and Malik~\cite{barron2014shape} and fine-tune FlowIID on the training set. All evaluation results are obtained using Euler’s method with a single time step.
\vspace{-0.5cm}
\begin{table}[t]
\centering
\caption{Quantitative comparison of Albedo on ARAP dataset. * implies model is finetuned on ARAP dataset}
\begin{tabular}{l c c c}
\hline
Method & LMSE$\downarrow$ & RMSE$\downarrow$ & SSIM$\uparrow$ \\
\hline
Niid-net*~\cite{luo2020niid}         & 0.023 & 0.129 & \textbf{0.788} \\
Lettry et al.~\cite{lettry2018unsupervised}      & 0.042 & 0.163 & 0.670 \\
Kocsis et al.~\cite{kocsis2024intrinsic}      & 0.030 & 0.160 & 0.738 \\
Zhu et al.~\cite{zhu2022learning}        & 0.029 & 0.184 & 0.729 \\
Intrinsicanything~\cite{chen2024intrinsicanything}        & 0.038 & 0.171 & 0.692 \\
Careaga and Aksoy~\cite{careagaIntrinsic}  & 0.025 & 0.140 & 0.671 \\
PIENet~\cite{das2022pie}              & 0.031 & 0.139 & 0.718 \\
Careaga and Aksoy ~\cite{careagaColorful}  & 0.023 & 0.145 & 0.700 \\
\textbf{Ours}           & \textbf{0.021} & \textbf{0.108} & 0.760 \\
\hline
\end{tabular}
\label{tab:alb_arap}
\end{table}
\begin{table}[h]
\centering
\caption{Quantitative comparison of Shading on ARAP Dataset. * implies model is finetuned on ARAP dataset}
\begin{tabular}{l c c c}
\hline
Method & LMSE$\downarrow$ & RMSE$\downarrow$ & SSIM$\uparrow$ \\
\hline
Niid-net*~\cite{luo2020niid}         & 0.022 & 0.206 & \textbf{0.781} \\
Lettry et al. ~\cite{lettry2018unsupervised}      & 0.042 & 0.193 & 0.610 \\
Careaga and Aksoy ~\cite{careagaIntrinsic}  & 0.026 & 0.168 & 0.680 \\
PIENet~\cite{das2022pie}              & 0.037 & 0.170 & 0.718 \\
\textbf{Ours}             & \textbf{0.022} & \textbf{0.132} & 0.744 \\
\hline
\end{tabular}
\label{tab:sh_arap}
\end{table}

\textbf{Performance Evaluation on MIT Intrinsic:}  
Table~\ref{tab:mit} reports quantitative results for albedo and shading prediction.  
FlowIID achieves the lowest LMSE for albedo (0.0043) and shading (0.0119), indicating strong consistency in structural reconstruction.  
While PIENet~\cite{das2022pie} attains the best albedo MSE and DSSIM, and PAIDNet~\cite{huang2025deep} excels in shading MSE and DSSIM, our method provides a balanced trade-off across metrics. Overall, these results show that FlowIID outperforms compared to existing methods.

\textbf{Performance Evaluation on ARAP:}
Tables~\ref{tab:alb_arap} and~\ref{tab:sh_arap} report our results on albedo and shading prediction. For albedo, our method achieves the lowest LMSE (0.021) and RMSE (0.108), outperforming all prior works, including ARAP-specific finetuned models.  
For shading, our model obtains the best RMSE (0.132) and competitive LMSE (0.022), while maintaining a strong SSIM (0.744). Although Niid-net~\cite{luo2020niid} reports a slightly higher SSIM, the model is finetuned on ARAP. It demonstrates that our model is based on strong generalization without dataset-specific tuning. 

\textbf{Comparison of Model Parameters:} Unlike recent diffusion-based IID approaches~\cite{rgbx} and convolutional neural network (CNN)-based methods~\cite{careagaIntrinsic,das2022pie,careagaColorful}, which typically require tens of iterative steps and rely on hundreds of millions of parameters, our model achieves comparable results with only 52M parameters in a single inference step, as shown in Table~\ref{tab:results}. This compact and streamlined design makes FlowIID substantially more efficient and practical for deployment in real-world vision pipelines.
\vspace{-0.5cm}
\begin{table}[ht!]
\centering
\caption{Comparison of Number of Parameters during inference. * implies parameters while training.}
\begin{tabular}{l c}
\hline
\textbf{Method} & \textbf{Parameters} \\
\hline
RGB $\leftrightarrow$ X ~\cite{rgbx} & 1.28 B \\
Niid-net ~\cite{luo2020niid}             &273.1 M \\
Careaga and Aksoy ~\cite{careagaIntrinsic}             & 252.05 M \\
Careaga and Aksoy ~\cite{careagaColorful}             & 548.18 M \\
PIENet ~\cite{das2022pie}                       &204.09 M\\
\textbf{Ours}                      & \textbf{51.71/58.36* M} \\
\hline
\end{tabular}
\label{tab:results}
\end{table}

\textbf{Ablation Study:}
We conduct ablation studies on the ARAP dataset to evaluate the impact of key architectural choices in the UNet (Tables~\ref{tab:comparison} and~\ref{tab:comparison_1}). Removing the concatenation layer results in a noticeable performance drop, highlighting its importance. Increasing the UNet depth from four to five modified residual blocks raises the parameter count by 7.6M but does not provide consistent gains. Our complete model, which employs four blocks with concatenation, achieves the best overall performance across all metrics for both albedo and shading. These findings validate that a compact architecture with selective components is sufficient to achieve strong performance.
\vspace{-0.5cm}
\begin{table}[h]
\centering
\caption{Ablation Studies on ARAP for Albedo component.}
\begin{tabular}{l c c c}
\hline
Method & LMSE$\downarrow$ & RMSE$\downarrow$ & SSIM$\uparrow$ \\
\hline
Without Concatenation      & 0.0242 & 0.121 & 0.744 \\
With five blocks     & 0.0223 & 0.112 & 0.755 \\
\textbf{Ours}             & \textbf{0.0205} & \textbf{0.108} & \textbf{0.760} \\
\hline
\end{tabular}
\label{tab:comparison}
\end{table}
\vspace{-1cm}
\begin{table}[h]
\centering
\caption{Ablation Studies on ARAP for Shading component.}
\begin{tabular}{l c c c}
\hline
Method & LMSE$\downarrow$ & RMSE$\downarrow$ & SSIM$\uparrow$ \\
\hline
Without Concatenation      & 0.0242 & 0.139 & 0.721 \\
With five blocks    & 0.0245 &  0.134 & 0.714 \\
\textbf{Ours}             & \textbf{0.0224} & \textbf{0.132} & \textbf{0.744} \\
\hline
\end{tabular}
\label{tab:comparison_1}
\end{table}
\section{Conclusion}
\label{sec:con}
We introduced FlowIID, a single-step latent flow–matching framework for IID. In contrast to diffusion-based methods that require multiple sampling steps, FlowIID achieves competitive performance in a single forward pass with less number of parameters, making it both memory and computation-efficient. Experiments on standard benchmarks confirm its effectiveness, and its lightweight nature highlights its potential for real-time vision applications. Future work will explore reducing the higher MSE observed near $t=0$ during training, and extending FlowIID to downstream tasks.
\bibliographystyle{IEEEbib}
\bibliography{strings,refs}

\end{document}